\documentclass[letterpaper, 10 pt, conference]{ieeeconf}  

\IEEEoverridecommandlockouts                              
\usepackage{graphicx}  
\usepackage{xcolor}
\usepackage{amsmath}  
\usepackage{cite}
\usepackage{algorithm}
\usepackage{algorithmic}
\usepackage{multirow}
\usepackage{float}
\let\labelindent\relax
\usepackage{enumitem}
\usepackage{hyperref}
\usepackage{booktabs} 
\usepackage{array}    
\usepackage{caption}  
\usepackage{color}
\usepackage{romannum}
\title{\LARGE \bf

FoAM: Foresight-Augmented Multi-Task Imitation Policy for \\ Robotic Manipulation
}

\author{Litao Liu$^{1}$, Wentao Wang$^{2}$, Yifan Han$^{3}$, Zhuoli Xie$^{1}$, Pengfei Yi$^{3}$, Junyan Li$^{3}$, Yi Qin$^{1}$, Wenzhao Lian$^{4*}$
\thanks{*Corresponding Author: Wenzhao Lian.} 
\thanks{Contact Wenzhao Lian (\texttt{lianwenzhao@gmail.com}) or Litao Liu (\texttt{liulitao6688@gmail.com}) for project details. This work was completed during the internship of Litao Liu at CoreNetic.ai.}
\thanks{$^{1}$CoreNetic.ai, $^{2}$University of Southern California, $^{3}$Institute of Automation, Chinese Academy of Sciences, $^{4}$School of Artificial Intelligence, Shanghai Jiao Tong University.} 
}

\begin{document}

\maketitle
\thispagestyle{empty}
\pagestyle{empty}

\begin{abstract}

Multi-task imitation learning (MTIL) has shown significant potential in robotic manipulation by enabling agents to perform various tasks using a single policy. This simplifies the policy deployment and enhances the agent’s adaptability across different scenarios. However, key challenges remain, such as maintaining action reliability (e.g., avoiding abnormal action sequences that deviate from nominal task trajectories) and generalizing to unseen tasks with a few expert demonstrations. To address these challenges, we introduce the Foresight-Augmented Manipulation Policy (FoAM), a novel MTIL policy that pioneers the use of multi-modal goal condition as input and introduces a foresight augmentation in addition to the general action reconstruction. FoAM enables the agent to reason about the visual consequences (states) of its actions and learn more expressive embedding that captures nuanced task variations. Extensive experiments on over 100 tasks in simulation and real-world settings demonstrate that FoAM significantly enhances MTIL policy performance, outperforming state-of-the-art baselines by up to 41\% in success rate. Meanwhile, we released our simulation suites, including a total of 10 scenarios and over 80 challenging tasks designed for manipulation policy training and evaluation. See the project homepage \textcolor{blue}{\href{https://projFoAM.github.io/}{https://projFoAM.github.io/}} for project details. 


\end{abstract}

\section{Introduction}

One of the main goals of robot learning is to develop a general-purpose agent capable of performing various tasks based on user commands. Multi-task imitation learning (MTIL) is a key approach that enables agents to learn multiple skills from expert demonstrations and trains an efficient policy, eliminating the need for complex, hardcoded solutions or reward functions. 
However, suboptimal policies and how to obtain generalization with limited training data are key issues that need to be addressed in MTIL policy development.
This requires the novel MTIL framework to enable the agent to learn task-independent skills and more expressive information from demonstrations, capturing task-specific details to ensure reliable execution of individual tasks and generalizing to unseen tasks and scenarios \cite{kroemer2020reviewrobotlearningmanipulation, zhu2018robot, rivero2023robotic}.

Previous research has shown that task adaptation in MTIL can be achieved by incorporating goal conditions into multi-task policy training \cite{fang2019survey, zhen20243d, brohan2022rt, brohan2023rt, kim2024openvla, bharadhwaj2024roboagent, haldar2024bakuefficienttransformermultitask, ding2019goal}. However, the low reliable execution of multi-task policies remains a challenge. Existing MTIL policies that align robotic actions with expert actions based on goal conditions often fail to reason about these distribution differences in the demonstrations of various tasks, severely impacting the agents’ performance on individual tasks. Meanwhile, most MTIL policies adopt uni-modal goal conditions \cite{haldar2024bakuefficienttransformermultitask, sundaresan2024rtsketchgoalconditionedimitationlearning, rivera2022visual, ni2024generate, jiang2022vima, yokoyama2024vlfm, nasiriany2019planning, bharadhwaj2024roboagent}, which results in some inherent limitations. For example, with limited demonstrations and without data augmentation, policies conditioned on task instructions have difficulty generalizing to unseen tasks (discuss in Section \textcolor{blue}{\ref{sec:experiment_result}}). While policies conditioned on goal images offer fine-grained guidance and even demonstrate striking zero-shot capabilities \cite{rivera2022visual}, they encounter ambiguities during task execution. Ambiguity means that the same goal image may exist for different tasks in the same scenario. For instance, when the robot was tasked with placing an object into a multi-layer locker, the resulting goal image was ambiguous due to closedness. The manipulated object in the initial image disappeared in the goal image, making it impossible to determine which specific layer the object was placed in just based on the goal image (see \textcolor{blue}{Figure~\ref{fig:real_env_snapshot}}).

In this paper, we introduce the Foresight-Augmented Manipulation Policy (FoAM), a novel MTIL policy designed to enhance the task performance of agents while addressing the limitations of uni-modal goal-conditioned policy. This approach is inspired by the perception ability of humans when performing tasks. When humans receive task instructions, they can easily foresee the state when completing the task in mind, and guide task execution based on this foresight and real-time actions until the expected results are achieved \cite{dezfouli2013actions}. Similarly, FoAM addresses the limitations of uni-modal goal-conditioned MTIL policies by utilizing a multi-modal goal condition (goal image and task instruction). It also innovatively attempts to leverage a fine-tuned vision-language model (VLM) \cite{brooks2023instructpix2pix} to autonomously generate goal images.  
During training, we apply an action loss to refine the policy’s behavior and novelly introduce a foresight loss to control the consequences of its actions. This allows the agent to reason about its action across diverse tasks, and handle the ambiguities and variations in expert demonstration data, leading to more forward-looking and precise action during inference. 
FoAM demonstrates significant effectiveness with evaluations across more than 100 tasks in both simulation and the real world. It outperforms state-of-the-art baselines, achieving an increase in success rate by up to 41\% in success rate. Our main contributions are summarized as follows: 

\begin{itemize}[topsep=0pt]
    \item We employ multi-modal goal condition to address the limitations of uni-modal goal-conditioned MTIL policies. FoAM demonstrates the capability to generalize to unseen tasks with limited expert demonstrations while ensuring efficient execution in ambiguous scenarios. Additionally, we attempt to involve a Vision-Language Model (VLM) in FoAM, which enables the agent to autonomously acquire goal images.
    \item We propose foresight augmentation, a novel module designed to enable agents to learn more expressive embedding and enhance task performance by aligning task instructions with the consequences of their actions. 
    \item We have open-sourced the first simulated dual-arm system with rich tasks (over 80) and scenarios (over 10). This system replicates the widely-used UR3e robot in MuJoCo \cite{todorov2012mujoco} and will be continuously maintained. It serves as a valuable simulation tool for developing robotic manipulation policies, such as MTIL and Sim2Real reinforcement learning policies \cite{lum2024dextrah}. 
\end{itemize}

\section{Related Work}
\textbf{Goal-conditioned Learning for Robotic Manipulation.} In recent years, significant progress has been made in single-task learning policies \cite{zhao2023learning, chi2023diffusion, buamanee2024bi, mishra2023generative, fu2024mobile}. However, to enable a wider adaptability, intelligent robots must be equipped with the ability to conduct diverse tasks and complete them effectively. Among current MTIL approaches, language-conditioned policies utilize large-scale datasets to achieve task generalization, or apply data augmentation techniques, such as vision generation models, to modify backgrounds and manipulated objects, enabling generalization across more tasks and scenarios with limited training data \cite{brohan2022rt, padalkar2023open, brohan2023rt, bharadhwaj2024roboagent, haldar2024bakuefficienttransformermultitask, ha2023scaling, reuss2023goal, kim2024openvla, hausman2017multi, shridhar2022cliport, brohan2023can, wang2024rise}. Despite the promising initial success, we found that language-conditioned policies often struggle with unseen tasks without sufficient data or extensive data augmentation. In parallel, some policies have introduced the goal image as task condition \cite{sundaresan2024rtsketchgoalconditionedimitationlearning, haldar2024bakuefficienttransformermultitask, rivera2022visual, mandlekar2023mimicgen, fang2024egocentric, black2023zero}. Compared to language inputs, images could provide more expressive representation at the pixel level, enabling stronger generalization capabilities, and even allowing agents to perform zero-shot tasks \cite{rivera2022visual}. However, goal images are susceptible to scenario ambiguity, where visually identical images can be produced by different tasks, causing actions inconsistent with intention. 

Collecting goal images requires human involvement, such as hand-drawing sketch \cite{sundaresan2024rtsketchgoalconditionedimitationlearning}, which reduces the autonomy of the agent.
Recent work has explored multi-modal goal conditions to enhance agent's task performance \cite{jiang2022vima, ni2024generate, reed2022generalist, alayrac2022flamingo, black2023zero}. These policies extract workspaces and manipulated objects from both task instruction and initial observation, then compose a multi-modal embedding based on predefined templates. In contrast, FoAM leverages a vision-language model to generate a goal image with semantic information. The generated goal image and task instruction are directly processed to infer actions and predict their consequences. By doing so, our method addresses the limitations of uni-modal goal-conditioned policies that rely solely on language instructions, goal images, or predefined templates.

\textbf{Agents with Vision Language Models.} 
In recent years, Vision-Language Models (VLMs) have been introduced to robotics \cite{brohan2023can, huang2022language, newcombe2015dynamicfusion, pang2022grasp, huang2023voxposer, qi2020learning, alayrac2022flamingo, yang2023dawn, zhi2024closed}, enabling more complex visual reasoning and multi-modal application.
Meanwhile, in the community of image editing, VLMs also demonstrated the ability to understand language instructions, editing real-world images and producing highly realistic visual effects\cite{zhang2024magicbrush, fang2024egocentric}. Works\cite{bharadhwaj2023visual, wang2018high, sridhar2024nomad} further validate that edited images can be interpreted by robotic agents, where the generated goal images are directly used as a goal condition. In contrast, we integrate VLMs seamlessly into our framework not only during the inference stage, but also in policy training. The generated goal images serve as the ``labels" to compute the reconstruction loss for the policy (Section\textcolor{blue}{~\ref{sec:vision_encouragement}}), coupling the action learning and action result prediction.

\section{METHOD}

We seek to develop an MTIL policy that can reliably generate actions consistent with semantics while effectively generalizing to unseen tasks with a small amount of training data, enabling agents to complete various tasks more efficiently and accurately. To achieve this, we introduce FoAM, a novel multi-modal goal-conditioned policy. In the following sections, we will provide an overview of the FoAM in \textcolor{blue}{\ref{sec:pipeline_overview}}, detail the fine-tuning of a cutting-edge visual-language model to generate goal images for FoAM in \textcolor{blue}{\ref{sec:fine_tune_ip2p}}, propose the foresight augmentation in \textcolor{blue}{\ref{sec:vision_encouragement}}, and  introduce the implementation details of FoAM in \textcolor{blue}{\ref{sec:implement_FoAM}}.

\begin{figure}[htbp]
\centering
\includegraphics[width=0.48\textwidth]{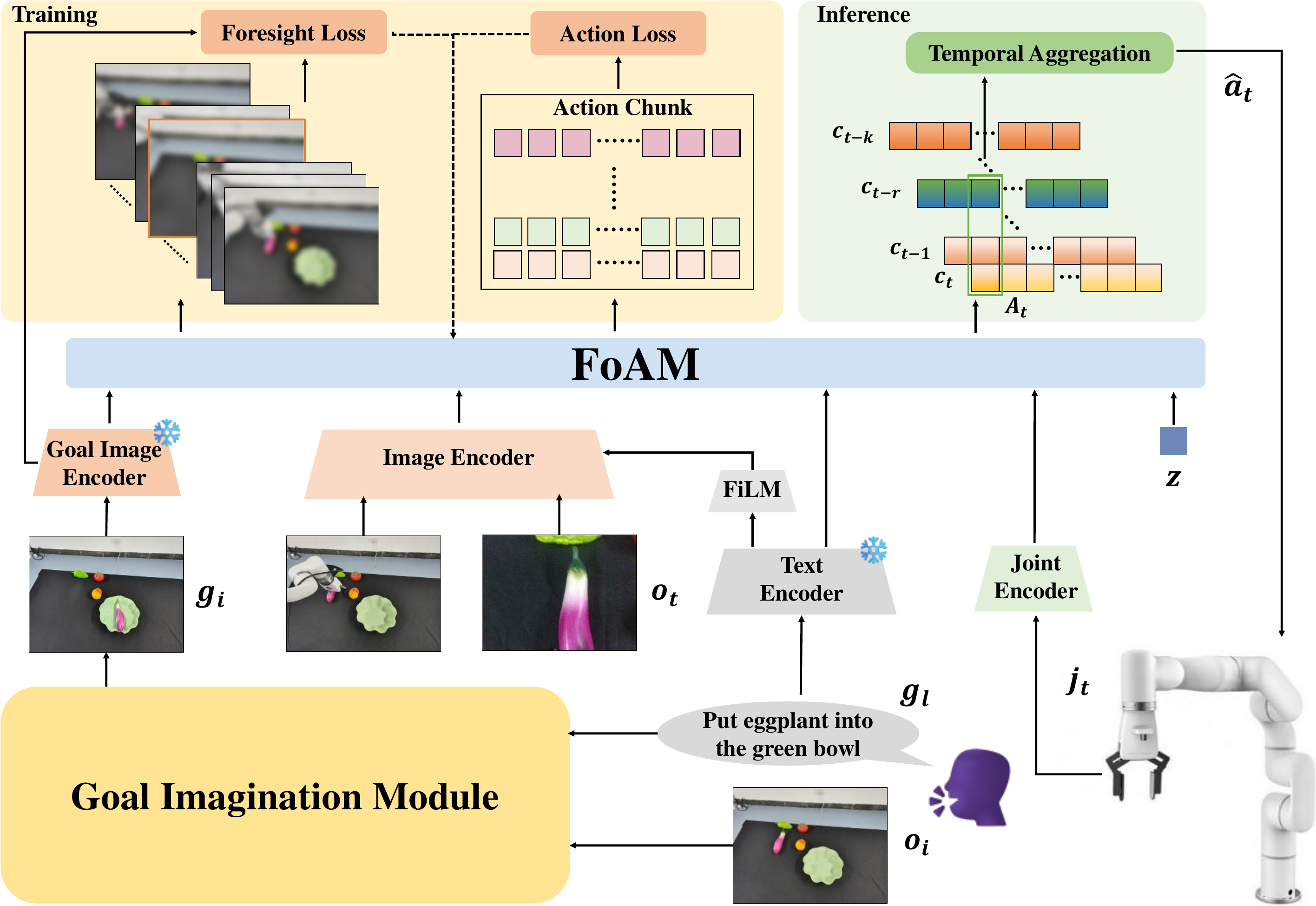}
\caption{\small \textit{Training and inference pipelines of FoAM.} The input terms remain consistent throughout both training and inference. During training, actions and their corresponding consequences are predicted, with the predicted consequences and actions being aligned with the input goal image and expert actions, respectively, to update the parameters of FoAM. During inference, the trained policy is used solely to predict the action $\hat{a}_{t}$.} 
\label{fig:pipeline}
\end{figure}

\subsection{Pipeline Overview}
\label{sec:pipeline_overview}

The pipeline of FoAM is illustrated in \textcolor{blue}{Figure \ref{fig:pipeline}}. FoAM is a transformer-based \cite{vaswani2017attention} policy that inherits the architecture of the prior work \cite{zhao2023learning} and is trained as a conditional variational autoencoder (CVAE) \cite{kingma2013auto, sohn2015learning}. The process begins with the user providing a task instruction \(g_l\) to the agent, which is then fed into FoAM as a language goal condition through a pre-trained text encoder \cite{gadre2023cows}. Concurrently, the task instruction, such as \textit{Put eggplant into the green bowl},  is input into the Goal Imagination Module along with the initial observation \(o_i\). The generated goal image or human-collected real goal image \(g_i\) serves as image goal condition. By integrating the agent's proprioception \(j_t\), visual observations \(o_t\), and the latent style variable \( z \) encoded from all episodes, the multi-modal goal-conditioned policy \(\pi_{\theta}(\hat{a}_{t:t+k} | o_t, j_t, g_i, g_l, z)\) is reconstructed by the coupling of foresight and action loss. During inference, action chunks are predicted and smooth actions are produced through temporal aggregation \cite{zhao2023learning}.

\subsection{Fine-tuned Goal Imagination Module}
\label{sec:fine_tune_ip2p}
The acquisition of goal images often requires human participation, such as hand-drawing sketch \cite{sundaresan2024rtsketchgoalconditionedimitationlearning} or collecting real goal images \cite{haldar2024bakuefficienttransformermultitask} before inference. In order to improve the efficiency of the agent's task execution, we embed a goal imagination module in FoAM to autonomously generate goal images. In this work, we chose InstructPix2Pix (Ip2p)\cite{brooks2023instructpix2pix} as the goal imagination module, leveraging a dataset of over 20,000 training pairs. Of these, 16,000 pairs were obtained from robot expert demonstrations provided by RT-1\cite{padalkar2023open, brohan2022rt, brohan2023rt}. The first and last frames of these demonstrations were used as the original and edited images, respectively, with the corresponding task name serving as the instruction. Since RT-1 demonstrations exhibited perturbations in the final frames due to robot arm movements, we implemented a rigorous data-cleaning procedure to remove noise and ensure high-quality training data. Additionally, we incorporated over 4,000 data pairs from our own simulation and real-world datasets (see Section \textcolor{blue}{\ref{sec:exp_dataset}} for details). 

\begin{figure}[htbp]
\centering
\includegraphics[width=0.5\textwidth]{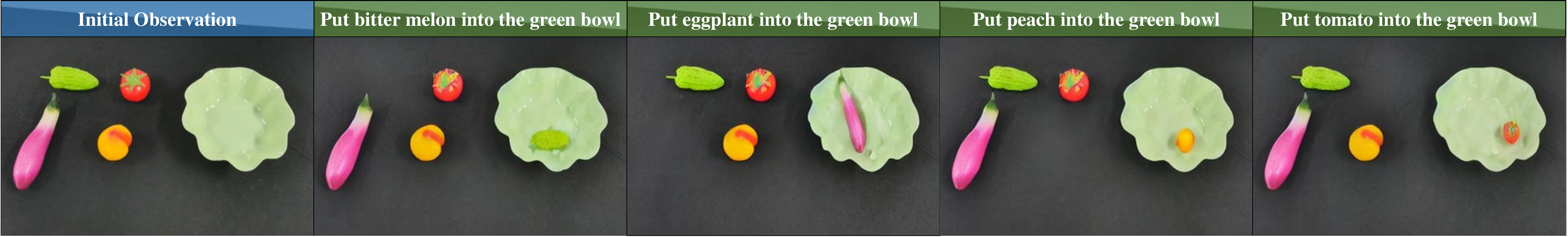}
\caption{\small \textit{Inference demonstrations of the fine-tuned goal imagination module.} The leftmost image illustrates the initial observation, while the next four images represent the edited goal images generated based on the initial observation and the task instruction provided at the top. Please visit the \textcolor{blue}{\href{https://projFoAM.github.io/}{project homepage}} for more examples.}
\label{fig:ip2p_results}
\end{figure}

We fine-tuned Ip2p for 500 epochs on a single NVIDIA H100 GPU, a process that required approximately 3 days. During the inference stage, with the model weights pre-loaded, generating one goal image of size $480\times640\times3$ took about 4 seconds. \textcolor{blue}{Figure \ref{fig:ip2p_results}} presents some inference demonstrations captured during VLM-FoAM joint inference experiments (Section \textcolor{blue}{\ref{sec:vlm_FoAM_joint_inference}}). They highlight the model’s ability to generate realistic and semantically consistent visuals based on the given initial observations and task instructions. 


\subsection{Foresight Augmentation}
\label{sec:vision_encouragement}
Humans possess exceptional perception abilities for understanding and interacting with external events. When performing a task, they can easily foresee the goal state of a scenario before execution in mind and use it as a guide.
Inspired by this capability, we developed a key module called Foresight Augmentation (FA) to equip agents with a similar perception mechanism. FA enables the agent to simultaneously comprehend both its actions and the subsequent consequences these actions will produce. By integrating FA into the agent’s decision-making process, we significantly enhance its overall performance in tasks execution (see Section \textcolor{blue}{\ref{sec:experiment_result}} for details).

We train FoAM as a CVAE, utilizing an encoder similar to those in \cite{zhao2023learning, bharadhwaj2024roboagent, shi2024yell, buamanee2024bi} to generate the latent variable $z\sim q_{\phi}(z|a_{t:t+k}, {j}_t)$. The decoder is defined as policy $\pi_{\theta}(\hat{a}_{t:t+k}, \hat{f}_{t:t+k}| o_t, j_t, g_i, g_l, z)$, which predicts a $k \times n$-dimensional action chunk $\hat{a}_{t:t+k}$ and foresight sequence $\hat{f}_{t:t+k}$ based on real-time observation and conditions, where $k$ represents the hyperparameter chunk size during training and $n$ denotes the dimension of the agent’s action space. To enhance the agent’s ability to interpret and respond to the dynamic work scenario, we strategically increase the value of $k$, thereby expanding the agent’s foresight horizon. For example, since each episode in the multi-task expert demonstration dataset $\mathcal{D}$ often have different time steps, we define the maximum time step of all episodes as $T$ and set $k$ to a value close to $T$. During training, while FoAM predicts the action chunk $\hat{a}_{t:t+k}$, FA generates $k$ potential foresight $\hat{f}_{t:t+k}$ and selects the frame $\hat{g}_{i} = \hat{f}_{t:t+k}[k-t]$ that is temporally consistent with the goal image $g_i$. This selected frame $\hat{g}_{i}$ is then aligned with the goal image $g_i$ using the foresight loss $L_\text{foresight}$ calculated by the Huber Loss, which is coupled with the action loss $L_\text{action}$ as the reconstruction loss $L_\text{recon}$ of the policy to play a role in updating the policy's parameters (\textcolor{blue}{Algorithm \ref{FoAM_training}}).
This process emulates the strong perception abilities humans exhibit when performing tasks, and we demonstrated in experiments that the FA significantly enhances the agent's task performance (Section \textcolor{blue}{\ref{sec:experiment_result}}).

\begin{algorithm}
\caption{FoAM Training}
\label{FoAM_training}  
\begin{algorithmic}[1]
\REQUIRE Expert demo $\mathcal{D}$, maximum episode time step $T$, chunk size $k$ ($k \approx T$), and loss weights $\alpha$, $\beta$, $\gamma$
\STATE Each episode includes $a_t$, $j_t$, $o_t$, $g_l$ and $g_i$, representing the action, agent proprioception, visual observation at time $t$, task prompt, and goal image, respectively. 
\STATE Init CVAE encoder $q_{\phi}(z|a_{t:t+k}, {j}_t)$
\STATE Init CVAE decoder $\pi_{\theta}(\hat{a}_{t:t+k}, \hat{f}_{t:t+k}| o_t, j_t, g_i, g_l, z)$
\FOR{each batch $i = 1, 2, \dots$}
\STATE Random sample $a_{t:t+k}$, $j_t$, $o_t$ from $\mathcal{D}$
\STATE Encode latent variable $z$ from $q_{\phi}(z|a_{t:t+k}, {j}_t)$
\STATE Predict actions $\hat{a}_{t:t+k}$ and foresight $\hat{f}_{t:t+k}$ using decoder $\pi_{\theta}(\hat{a}_{t:t+k}, \hat{f}_{t:t+k}| o_t, j_t, g_i, g_l, z)$
\STATE $\mathcal{L}_{\text{action}} = \text{$L_1$}(\hat{a}_{t:t+k}, a_{t:t+k})$
\STATE $\mathcal{L}_{\text{foresight}} = \text{$Huber$}(\hat{g}_{i}, g_i)$, where $\hat{g}_{i} = \hat{f}_{t:t+k}[k-t]$
\STATE $\mathcal{L}_{\text{reg}} = D_{\text{KL}}(q_{\phi}(z|a_{t:t+k}, {j}_t) \parallel \mathcal{N}(0, I))$
\STATE Update CVAE parameters $\theta$ and $\phi$ using ADAM optimizer with total loss $\mathcal{L} = \alpha\mathcal{L}_{\text{action}} + \beta\mathcal{L}_{\text{foresight}} + \gamma \mathcal{L}_{\text{reg}}$
\ENDFOR
\end{algorithmic}
\end{algorithm}

\subsection{FoAM Policy Implementation}
\label{sec:implement_FoAM}

FoAM is designed as a transformer-based policy with sufficient capacity to predict specific sequences by effectively integrating sequence information from the inputs. FoAM is implemented using an ACT-like architecture \cite{zhao2023learning} with the CVAE framework. 
The language-conditioned embedding is obtained by the pre-trained language encoder \cite{gadre2023cows} to produce a 384-dimensional feature, which is subsequently projected to 512 dimensions through an MLP. ResNet18 \cite{he2016deep} with FiLM conditional layer \cite{perez2018film} is used to encode visual observations of size $480\times640\times3$ and embed task instructions into images, ensuring robust task performance in multiple scenarios \cite{bharadhwaj2024roboagent}. The visual observations are finally transformed into a $(300 \times n)\times512$ feature sequence, where $n$ denotes the number of used viewpoints. The goal image $g_i$ is encoded by the pre-trained ResNet18, producing a $300\times512$ feature, and remains fixed during training without parameter updates. The latent variable $z$ is obtained with a 4-layer transformer encoder and projected to 512 dimensions. Proprioceptive input $j_t$ is projected to 512 dimensions through an MLP. The CVAE decoder consists of a 4-layer transformer encoder and a 7-layer transformer decoder. The input feature dimensions for the transformer encoder are $(303+300\times n)\times512$. The encoder fuses features from different modalities, and the decoder predicts the action chunk $\hat{a}_{t:t+k}$, while generating $k$ foresight $\hat{f}_{t:t+k}$ (each $300\times512$) through a second-dimensional full connection layer.

The FoAM training process, which incorporates the FA, is outlined in \textcolor{blue}{Algorithm \ref{FoAM_training}}. During training, we use L1 Loss and Huber Loss to compute the action loss $\mathcal{L}_{\text{action}}$ and foresight loss $\mathcal{L}_{\text{foresight}}$ respectively, along with a KL divergence term $\mathcal{L}_{\text{reg}}$ regularizing the CVAE encoder. These losses are weighted by $\alpha$, $\beta$, and $\gamma$. In our experiments, the weight values are set to $1$, $2$, and $10$, respectively.

During inference, the FA module is discarded. The implemented policy is represented by \(\pi_{\theta}(\hat{a}_{t:t+k} | o_t, j_t, g_i, g_l, z)\). Based on the current observations and goals, the action chunk $c_t=\hat{a}_{t:t+k}$ is predicted. Following prior action chunk-based policies \cite{bharadhwaj2024roboagent, zhao2023learning, haldar2024bakuefficienttransformermultitask}, we apply exponential temporal aggregation to produce smooth action trajectories. Unlike previous work, we introduce the hyperparameter temporal aggregation range $r$ as the actual chunk size during inference, which eliminates the equality constraint on chunk size $k$ during training and inference. This is particularly crucial for deploying FoAM, as it allows flexibly adjusting the aggregation range according to the user’s intentions and the characteristics of different tasks, and optimizing task performance. During experiments, we observed that when $r$ is large ($>100$), the agent can perform tasks more efficient, and when $r$ is relatively small ($50\sim80$), the agent has better reactiveness and ability to resist external disturbance (Section \textcolor{blue}{\ref{sec:robustness_analysis}}).
The inference code is shown in \textcolor{blue}{Algorithm \ref{FoAM_inference}}. There are approximately $160M$ parameters in the training and around $80M$ in the inference.

\begin{algorithm}
\caption{FoAM Inference}
\label{FoAM_inference}  
\begin{algorithmic}[1]
\REQUIRE trained policy \(\pi_{\theta}(\hat{a}_{t:t+k} | o_t, j_t, g_i, g_l, z)\), where $z=0$, maximum inference time step $L$, chunk size $k$, temporal aggregation range $r$ and weight coefficient $\lambda$. 
\STATE Init an action buffer $B[L, L+k, *]$, where $B[t]$ stores action chunk $\hat{a}_{t:t+k}$.
\FOR{time step $t$ = range$(L)$}
\STATE Predict $\hat{a}_{t:t+k}$ with \(\pi_{\theta}\)
\STATE Add $\hat{a}_{t:t+k}$ to buffer $B[t, t:t+k]$ 
\STATE Extract temporal aggregation array $A_t = B[-r:, t]$ 
\STATE Get $\hat{a}_t = \sum_{i} w_i A_t[i] / \sum_{i} w_i$, with $w_i = \exp(-\lambda * i)$
\ENDFOR
\end{algorithmic}
\end{algorithm}

\section{Experiments}
\label{sec:EXPERIMENTS}
Our experiments will answer the following questions:
\begin{enumerate}[label=(\alph*)]
    

    \item How does multi-modal goal condition perform?
    \item How does foresight augmentation perform?
    \item Can FoAM generalize to unseen tasks with a few expert episodes and without data augmentation?
    \item How does FoAM perform differently when guided by a real or generated goal image?
    \item How well does FoAM respond to external disturbance? 
    
\end{enumerate}

\begin{table*}[ht]
    \centering
    \begin{tabular}{p{2cm}p{1cm}|p{1.5cm}p{1.5cm}p{1.8cm}p{1.8cm}p{1.1cm}p{1.7cm}}
    \toprule
    \textbf{Policy} & \textbf{Model size} & \textbf{Dual-Arm (12 tasks)} & \textbf{Block-based (40 tasks)} & \textbf{Cabinet-based (14 tasks)} & \textbf{Locker-based (8 tasks)} & \textbf{Others (8 tasks)} & \textbf{Unseen Tasks (4 tasks)} \\
    \midrule
    BAKU\cite{haldar2024bakuefficienttransformermultitask}  &11M&31\% & 47\% & 52\% & 25\% & 17\% & 0 \\
    MT-ACT\cite{bharadhwaj2024roboagent} &86M&33\% &	71\% & 50\% & 32\% & 50\% & 0 \\
    Gimg-ACT  &84M&45\% & 39\% & 52\% & 23\% & 28\% & 45\% \\
    FoAM \textbf{(Ours)}  &86M&\textbf{86\%} & \textbf{91\%} & \textbf{75\%} & \textbf{85\%} & \textbf{71\%} & \textbf{66\%} \\
    FoAM w/o FA  &86M&36\% & 81\% & 55\% & 51\% & 49\% & 11\% \\
    \bottomrule
    \end{tabular}
    \caption{\small \textit{Performance of interested policies in FoAM benchmark.} In the table, the first column lists the names of the policies included in the evaluation, and the second column provides the model size of each policy. The subsequent six columns report the average success rates of the respective policies across five task categories and four unseen tasks.}  
    \label{tab:sim_result_table}
\end{table*}

\begin{table}[ht]
    \raggedright
    \centering
    \begin{tabular}{p{1.7cm}|p{0.9cm}p{0.9cm}p{0.9cm}p{0.9cm}p{0.9cm}}
    \toprule
    \textbf{Policy} & \textbf{Scenario I} & \textbf{Scenario II} & \textbf{Scenario III} & \textbf{Scenario IV} & \textbf{Unseen Task}\\
    \midrule
    MT-ACT                  
        &6/40&10/40  &	10/40 & 10/20 & 0/10\\
    Gimg-ACT                
        &7/40&8/40  &	11/40 & - & 1/10\\

    FoAM \textbf{(Ours)}    
        &\textbf{11/40}&11/40 & \textbf{17/40} & \textbf{14/20} & \textbf{3/10}\\
    FoAM w/o FA            
    &7/40&\textbf{13/40}  &	13/40 & 12/20 & 1/10\\
    \bottomrule
    \end{tabular}
    \caption{\small \textit{Performance of interested policies in real-world scenarios.} Scenario I corresponds to the task: ``Pick the \textit{first (second, third, forth)} test tube from the rack.'' Scenario II corresponds to: ``Insert the test tube into the \textit{first (second, third, forth)} hole.'' Scenario III involves: ``Put the \textit{eggplant (bitter melon, peach, tomato)} in the green bowl.'' Scenario IV refers to the task: ``Place the bitter melon on the \textit{bottom (middle)} locker layer.''
}  
    \label{tab:real_exp_results}
\end{table}

\subsection{Data Collection}
\label{sec:exp_dataset}
\begin{figure}[htbp]
\centering
\includegraphics[width=0.48\textwidth]{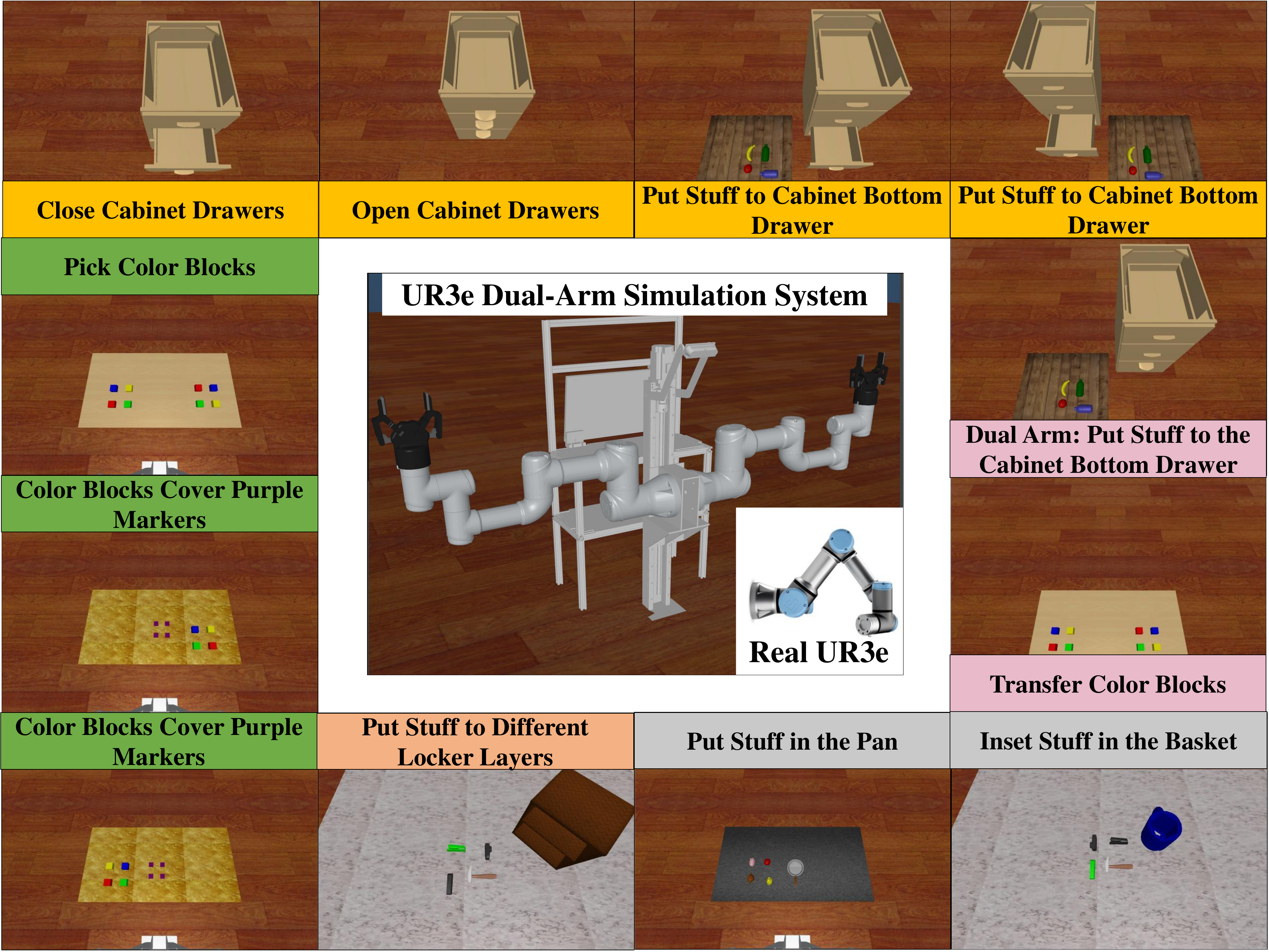}
\caption{\small \textit{Snapshots of each scenario in the FoAM benchmark.} The middle snapshot provides an overview of the simulated dual-arm system we developed in MuJoCo \cite{todorov2012mujoco}. The tasks in the benchmark are divided into five categories for evaluation: pink for dual-arm tasks, yellow for cabinet-based tasks, green for color-block-based tasks, orange for locker-based tasks, and gray for other tasks. The objects in these scenarios are sourced from \cite{dasari2023pgdm, Xiang_2020_SAPIEN, Mo_2019_CVPR, chang2015shapenet}. The FoAM benchmark offers high-degree-of-freedom simulation suites. Tutorials for creating custom environments are available on the \textcolor{blue}{\href{https://projFoAM.github.io/}{project homepage}}.}
\label{fig:sim_env_snapshot}
\end{figure}

{\textbf{FoAM Benchmark}}. We developed a dual-arm system in MuJoCo\cite{todorov2012mujoco}, a popular physics simulation engine, with 6 degrees of freedom (DoF) each arm and a 1-DOF parallel-jaw gripper. This system is the first open-source simulated dual-arm system with rich tasks, manipulation scenarios, and the same physics as the real UR3e robot. It will serve as a valuable tool for developing zero-shot Sim2Real robotic reinforcement learning policies, such as DextrAH-G\cite{lum2024dextrah}. We have designed 10 distinct multi-task suites in the system. A total of 86 simulation tasks were involved, encompassing a broad range of practical skills, such as picking, moving, pushing, placing, sliding, inserting, opening, closing, and transferring \cite{padalkar2023open}. \textcolor{blue}{Figure \ref{fig:sim_env_snapshot}} provides an overview of snapshots from each multi-task scenario along with their corresponding names. Each scenario includes varying numbers of subtasks. For example, the Open Cabinet Drawer scenario consists of three subtasks, with a general task instruction “Open the cabinet bottom drawer”, where “bottom” can be replaced with “middle” or “top”. The subtasks in the scenarios Transfer Color Blocks and Dual Arm: Put Stuff to the Cabinet Bottom Drawer are dual-arm tasks, while the remaining suites involve single-arm tasks. The FoAM benchmark is a high-degree-of-freedom simulation data generator, enabling users to customize textures, colors, and even trajectories. This tool facilitates the rapid generation of high-quality simulation data tailored to user requirements by running scripts. The FoAM benchmark will enrich the existing simulation benchmark libraries, such as RLBench\cite{james2020rlbench} and LIBERO \cite{ liu2023libero}. We expect that FoAM will contribute to the development of multi-task policies in complex scenarios.

{\textbf{Simulation Dataset}}. We generate 50 episodes for each task. Before recording each demonstration, objects in the scene are randomly initialized within a specified range. The dataset is recorded at a frequency of 50 Hz, capturing the robot's proprioceptive data, action sequences, and visual observations. The visual observations are captured solely through a head-mounted camera with a resolution of $480\times640$ pixels. For the single-arm tasks, even though one of the robotic arms remained inactive, we still recorded its action and proprioceptive data. As a result, the controlled action space $n$ of the dataset is unified to $14$, allowing us to accommodate all tasks within a single MTIL policy.

\begin{figure}[htbp]
\centering
\includegraphics[width=0.48\textwidth]{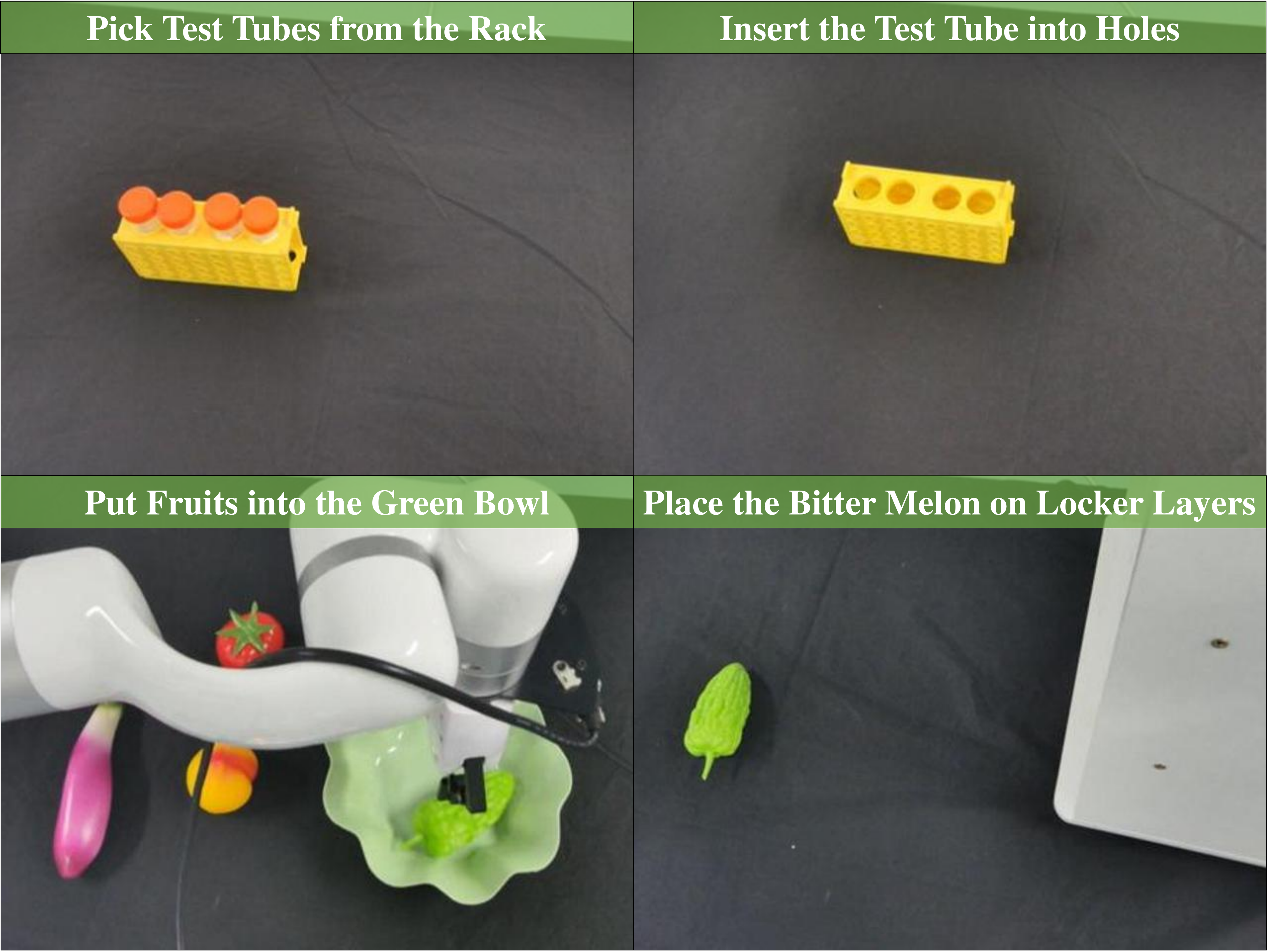}
\caption{\small \textit{Snapshots of the real-world multi-task environment are captured from static externally mounted Orbbec Femto Bolt camera.}  
The tasks include \textit{Pick test tubes from the rack}, \textit{Put fruits into the green bowl}, \textit{Insert the test tube into holes} (four subtasks each), and \textit{Place the bitter melon on locker layers} (two subtasks). Please refer to the \textcolor{blue}{\href{https://projFoAM.github.io/}{project homepage}} for videos and more details.} 
\label{fig:real_env_snapshot}
\end{figure}

{\textbf{Real World Dataset}}. Our robot system is composed of a UFACTORY xArm 7 robotic arm, a parallel-jaw gripper, a static externally mounted Orbbec Femto Bolt camera, and a wrist-mounted Intel Realsense D435 camera. To evaluate the performance of FoAM in the real world, we designed 14 tasks across four multi-task scenarios in the real world. The snapshots of four scenarios are illustrated in \textcolor{blue}{Figure \ref{fig:real_env_snapshot}}. The dataset was collected using a custom-built, low-cost teleoperation platform inspired by Gello~\cite{wu2024gellogenerallowcostintuitive}. 
We collected 50 episodes for each task, with the objects randomly placed on the table before data collection. The randomization was constrained within a rectangular area measuring approximately 50$\times$60 cm. 
The final dataset comprises RGB data from two cameras with a resolution of 480$\times$640 each, along with joint states from both the leading and following arms, recorded at a frequency of 30 Hz. 

\subsection{Experiment Results} 
\label{sec:experiment_result}

We compare FoAM against state-of-the-art open-source MTIL policies, including \textit{Multi-task Action Chunking Transformer (MT-ACT)}   \cite{bharadhwaj2024roboagent}  and  \textit{BAKU with a deterministic policy head}  \cite{haldar2024bakuefficienttransformermultitask},  both of which utilize only task instruction as the goal condition.  We evaluate  \textit{ACT  with goal images  (Gimg-ACT)}  as a baseline guided solely by the goal image.  The baseline of \textit{ACT  with both task instruction and goal image} serves as the multi-modal goal-conditioned policy, which can also function as an ablation experiment to assess the effectiveness of the foresight augmentation \textit{(FoMA w/o FA)}.

All scenarios in the FoAM benchmark were categorized into five distinct task categories and four unseen tasks. We conducted 50 test trials for each task, and the average success rates of the different policies across these categories are presented in \textcolor{blue}{Table \ref{tab:sim_result_table}}. Compared to all the policies evaluated, FoAM achieved the highest success rate across all task categories. Notably, in dual-arm tasks, FoAM outperformed the second-best policy (Gimg-ACT)  by  41\%  in success rate,  with varying degrees of improvement observed in the other task categories as well.

{\textbf{Advantages of multi-modal goal condition}}. The multi-modal goal condition enables the agent to perform unseen tasks with limited expert demonstrations and without data augmentation while solving the scenario ambiguity involved in the goal-image-based uni-modal policy. To evaluate the generalization capabilities of interested policies,  we designed  four  unseen tasks by modifying the Scenario \textit{Pick Color Blocks}: green blocks were changed to purple,  and blue blocks to black. The language-based uni-modal policies (MT-ACT, BAKU) could not complete these unseen tasks, while policies involving goal image (Gimg-ACT, FoAM, FoAM w/o FA) demonstrated varying levels of generalization. We attribute this to the language-based policies that rely on text embeddings to conduct conditional responses, executing tasks based on the broad sub-tasks level. These policies leave the agent struggling when faced with unseen task instructions. In contrast, image-based uni-modal policies take actions based on pixel information, aligning goal information with a single episode, which would provide a more fine-grained representation on the episode level. This enables the policy to focus on precise pixel differences between visual observations and goals, allowing it to tackle unseen tasks.

The goal-image-based uni-modal policy suffers from scenario ambiguity, making it difficult to align tasks with intentions. For example, the scenarios in Block-based category, such as \textit{Transfer Color Blocks} and \textit{Pick Color Blocks}, have the same initial observation and goal images. The tasks, such as \textit{Transfer the right yellow block} and \textit{Pick the right yellow block}, have the same initial observation and goal image, which is the visual observation after the right yellow block disappears in the filed of view. Therefore, Gimg-ACT has the worst performance in Block-based tasks compared to other policies.  The multi-modal goal condition of FoAM provides reliable execution conditions for each task while retaining the generalization ability offered by goal images. 

{\textbf{Advantages of Foresight Augmentation}}. To discuss the benefits brought by Foresight Augmentation (FA), we designed the ablation item FoAM w/o FA. A comparative analysis of success rates across all five task categories reveals a significant performance enhancement in task execution after integrating the FA. We attribute this improvement to the policy's ability to learn more expressive representations by aligning actions with their consequences, thereby facilitating more accurate task execution. 
According to \textcolor{blue}{Table \ref{tab:sim_result_table}}, our experiment results indicate that the ability to perform unseen tasks follows the order FoAM$>$Gimg-ACT$>$FoAM w/o FA, while language-based goal-conditioned policies (MT-ACT and BAKU) fail to generalize to unseen tasks. These results suggest that incorporating task instructions into the multi-modal goal condition influences the generalization capability conferred by the goal image. However, the integration of FA mitigates the negative impact of task instructions on generalization while enhancing the acquisition of more expressive representations, ultimately improving the policy’s overall generalization performance.

\begin{table}[ht]
    \raggedright
    \centering
    \begin{tabular}{p{1.6cm}|p{1.6cm}p{1.2cm}p{1cm}p{1cm}}
    \toprule
    \textbf{Policy} & \textbf{Bitter Melon} & \textbf{Eggplant} & \textbf{Peach} & \textbf{Tomato} \\
    \midrule
    FoAM                  
        &5/10& 7/10  &	2/10 & 2/10\\
    VLM-FoAM                
        &7/10& 7/10  &	3/10 & 3/10 \\
    \bottomrule
    \end{tabular}
    \caption{\small \textit{Performance comparison of the FoAM and VLM-FoAM policies in Scenario III.} The first column lists the evaluated policies, while the last four columns present the success rates for operating each corresponding object in Scenario III.}
    \label{tab:vlm_foam_exp_results}
\end{table}

{\textbf{Real-world experiment}}. Based on the performance of interested policies in FoAM benchmark, we strategically selected MT-ACT, Gimg-ACT, FoAM, and FoAM w/o FA for real-world deployment. In the real-world experiments, for each scenario, we randomly initialized ten different locations and sequentially executed the tasks associated with each scenario. The performance of these MTIL policies in real-world scenarios are summarized in \textcolor{blue}{Table \ref{tab:real_exp_results}}. 

Each policy experienced a notable performance decline when deployed in real-world environments. We attribute this to the inherent randomness of expert demonstrations,  distribution differences in multiple tasks, and the variability present in the dynamic real world, all of which complicate the processes of learning and inference. 
Furthermore, we utilized only two perspectives, head and wrist, and the manipulated objects were randomly initialized within a large workspace, increasing the challenge for the agent to learn skills effectively. Scenarios I and II demand higher accuracy from the actions, as the test tubes and racket holes are closely located, making them sensitive to robot misoperations. In Scenario III, the four fruits and the green bowl were placed randomly, resulting in complex visual observations. In contrast, Scenario IV was relatively simple for robot execution, involving just one large manipulated object and a well-defined goal space. However, Scenario IV introduced ambiguity, making it difficult for Gimg-ACT to reliably execute tasks consistent with intention. 

Consistent with the simulation results, FoAM demonstrates superior performance across nearly all four real world scenarios. To evaluate policy performance on unseen tasks, we replaced the eggplant in Scenario III with a carambola. MT-ACT still struggled to achieve any success in the new tasks. The other three policies exhibited varying degrees of generalization, and FoAM achieved the highest success rate.


\subsection{VLM-FoAM Joint Inference}
\label{sec:vlm_FoAM_joint_inference}
To improve the agent's autonomy in acquiring the goal image, we conducted a joint inference experiment in Scenario III. Two policies were employed using data exclusively from Scenario III: FoAM, trained with the last frame of demonstration and evaluated with real goal images, and VLM-FoAM, trained and evaluated with goal images generated by VLM. The experiment results are shown in \textcolor{blue}{Table \ref{tab:vlm_foam_exp_results}}. 

Due to their shapes, Peach and Tomato are prone to rolling, which is difficult for the robot to grasp, leading to task failure. In contrast, Bitter Melon and Eggplant are more easily grasped. Although the experimental results have a small gap, we observed that VLM-FoAM exhibits more robust performance during the experiments. We attribute this to the deep semantic information retained in the images generated by VLM, which helps prevent the model from overfitting when working with small datasets. Furthermore, the goal images generated by VLM maintain a consistent overall style. This style uniformity ensures that goal images generated at different times share similar features, enhancing the robot's ability to adapt to the dynamic real world, thereby improving task execution reliability. Additionally, with the introduction of VLM, the agent can autonomously and efficiently acquire the goal image, with a 480$\times$640 pixel goal image being obtained in an average of 4 seconds.

Although VLM brings benefits to robotic manipulation, we found some limitations during the experiment. When executing tasks in scenarios with complex workspaces and small manipulated objects, such as Scenario I and Scenario II, the generated goal images exhibited instability in semantic alignment. For example, when given the instruction \textit{Insert the test tube into the second hole}, the generated image sometimes corresponded to an incorrect but visually similar placement, such as \textit{Inserting the test tube into the first/third hole}. We attribute these discrepancies to the limited training data and the limitations of current VLMs in processing fine-grained pixel-level details. For instance, the test tube's features constitute only a small portion of the overall visual input, posing a significant challenge for the model in accurately capturing and distinguishing such fine-grained spatial relationships.

\subsection{Robustness Analysis}
\label{sec:robustness_analysis}
We conducted an in-depth exploration of FoAM, focusing on two key aspects: external disturbance, and reactiveness. Relevant videos can be viewed on the \textcolor{blue}{\href{https://projFoAM.github.io/}{project homepage}}.

\textbf{External Disturbance.} 
Despite the introduction of additional objects to disrupt the operation process, the robot was able to complete the task without significant difficulties.

\textbf{Reactiveness.}
During the task execution, we forcibly removed the object from the gripper. In response, the robot exhibited the ability to attempt re-grasping the object and ultimately complete the task.

\section{Conclusion and Future Work}
In this work, we introduced FoAM, a novel multi-modal goal-conditioned policy designed to enhance the performance of multi-task policies and address the limitations of uni-modal ones. Inspired by human action perception, FOAM improves the agent’s performance by mimicking expert actions while aligning actions with their consequences. In our published FoAM benchmark and across real-world scenarios, FoAM achieved improvements of up to 41\% in success rate compared with previous policies. However, FoAM exhibited certain limitations in real-world Scenarios I and II, which involve high precision requirements. To address this, we will explore refining long-horizon tasks by generating fine-grained intermediate goal images to serve as consequence controllers. By leveraging them, we seek to reduce cumulative errors during manipulations and improve the agent’s execution accuracy.



\bibliographystyle{IEEEtran}
\bibliography{ref}

\begin{thebibliography}{10}
\providecommand{\url}[1]{#1}
\csname url@samestyle\endcsname
\providecommand{\newblock}{\relax}
\providecommand{\bibinfo}[2]{#2}
\providecommand{\BIBentrySTDinterwordspacing}{\spaceskip=0pt\relax}
\providecommand{\BIBentryALTinterwordstretchfactor}{4}
\providecommand{\BIBentryALTinterwordspacing}{\spaceskip=\fontdimen2\font plus
\BIBentryALTinterwordstretchfactor\fontdimen3\font minus \fontdimen4\font\relax}
\providecommand{\BIBforeignlanguage}[2]{{%
\expandafter\ifx\csname l@#1\endcsname\relax
\typeout{** WARNING: IEEEtran.bst: No hyphenation pattern has been}%
\typeout{** loaded for the language `#1'. Using the pattern for}%
\typeout{** the default language instead.}%
\else
\language=\csname l@#1\endcsname
\fi
#2}}
\providecommand{\BIBdecl}{\relax}
\BIBdecl

\bibitem{kroemer2020reviewrobotlearningmanipulation}
\BIBentryALTinterwordspacing
O.~Kroemer, S.~Niekum, and G.~Konidaris, ``A review of robot learning for manipulation: Challenges, representations, and algorithms,'' 2020. [Online]. Available: \url{https://arxiv.org/abs/1907.03146}
\BIBentrySTDinterwordspacing

\bibitem{zhu2018robot}
Z.~Zhu and H.~Hu, ``Robot learning from demonstration in robotic assembly: A survey,'' \emph{Robotics}, vol.~7, no.~2, p.~17, 2018.

\bibitem{rivero2023robotic}
Y.~Rivero-Moreno, S.~Echevarria, C.~Vidal-Valderrama, L.~Pianetti, J.~Cordova-Guilarte, J.~Navarro-Gonzalez, J.~Acevedo-Rodr{\'\i}guez, G.~Dorado-Avila, L.~Osorio-Romero, C.~Chavez-Campos \emph{et~al.}, ``Robotic surgery: a comprehensive review of the literature and current trends,'' \emph{Cureus}, vol.~15, no.~7, 2023.

\bibitem{fang2019survey}
B.~Fang, S.~Jia, D.~Guo, M.~Xu, S.~Wen, and F.~Sun, ``Survey of imitation learning for robotic manipulation,'' \emph{International Journal of Intelligent Robotics and Applications}, vol.~3, pp. 362--369, 2019.

\bibitem{zhen20243d}
H.~Zhen, X.~Qiu, P.~Chen, J.~Yang, X.~Yan, Y.~Du, Y.~Hong, and C.~Gan, ``3d-vla: A 3d vision-language-action generative world model,'' \emph{arXiv preprint arXiv:2403.09631}, 2024.

\bibitem{brohan2022rt}
A.~Brohan, N.~Brown, J.~Carbajal, Y.~Chebotar, J.~Dabis, C.~Finn, K.~Gopalakrishnan, K.~Hausman, A.~Herzog, J.~Hsu \emph{et~al.}, ``Rt-1: Robotics transformer for real-world control at scale,'' \emph{arXiv preprint arXiv:2212.06817}, 2022.

\bibitem{brohan2023rt}
A.~Brohan, N.~Brown, J.~Carbajal, Y.~Chebotar, X.~Chen, K.~Choromanski, T.~Ding, D.~Driess, A.~Dubey, C.~Finn \emph{et~al.}, ``Rt-2: Vision-language-action models transfer web knowledge to robotic control,'' \emph{arXiv preprint arXiv:2307.15818}, 2023.

\bibitem{kim2024openvla}
M.~J. Kim, K.~Pertsch, S.~Karamcheti, T.~Xiao, A.~Balakrishna, S.~Nair, R.~Rafailov, E.~Foster, G.~Lam, P.~Sanketi \emph{et~al.}, ``Openvla: An open-source vision-language-action model,'' \emph{arXiv preprint arXiv:2406.09246}, 2024.

\bibitem{bharadhwaj2024roboagent}
H.~Bharadhwaj, J.~Vakil, M.~Sharma, A.~Gupta, S.~Tulsiani, and V.~Kumar, ``Roboagent: Generalization and efficiency in robot manipulation via semantic augmentations and action chunking,'' in \emph{2024 IEEE International Conference on Robotics and Automation (ICRA)}.\hskip 1em plus 0.5em minus 0.4em\relax IEEE, 2024, pp. 4788--4795.

\bibitem{haldar2024bakuefficienttransformermultitask}
\BIBentryALTinterwordspacing
S.~Haldar, Z.~Peng, and L.~Pinto, ``Baku: An efficient transformer for multi-task policy learning,'' 2024. [Online]. Available: \url{https://arxiv.org/abs/2406.07539}
\BIBentrySTDinterwordspacing

\bibitem{ding2019goal}
Y.~Ding, C.~Florensa, P.~Abbeel, and M.~Phielipp, ``Goal-conditioned imitation learning,'' \emph{Advances in neural information processing systems}, vol.~32, 2019.

\bibitem{sundaresan2024rtsketchgoalconditionedimitationlearning}
\BIBentryALTinterwordspacing
P.~Sundaresan, Q.~Vuong, J.~Gu, P.~Xu, T.~Xiao, S.~Kirmani, T.~Yu, M.~Stark, A.~Jain, K.~Hausman, D.~Sadigh, J.~Bohg, and S.~Schaal, ``Rt-sketch: Goal-conditioned imitation learning from hand-drawn sketches,'' 2024. [Online]. Available: \url{https://arxiv.org/abs/2403.02709}
\BIBentrySTDinterwordspacing

\bibitem{rivera2022visual}
C.~G. Rivera, D.~A. Handelman, C.~R. Ratto, D.~Patrone, and B.~L. Paulhamus, ``Visual goal-directed meta-imitation learning,'' in \emph{Proceedings of the IEEE/CVF Conference on Computer Vision and Pattern Recognition}, 2022, pp. 3767--3773.

\bibitem{ni2024generate}
F.~Ni, J.~Hao, S.~Wu, L.~Kou, J.~Liu, Y.~Zheng, B.~Wang, and Y.~Zhuang, ``Generate subgoal images before act: Unlocking the chain-of-thought reasoning in diffusion model for robot manipulation with multimodal prompts,'' in \emph{Proceedings of the IEEE/CVF Conference on Computer Vision and Pattern Recognition}, 2024, pp. 13\,991--14\,000.

\bibitem{jiang2022vima}
Y.~Jiang, A.~Gupta, Z.~Zhang, G.~Wang, Y.~Dou, Y.~Chen, L.~Fei-Fei, A.~Anandkumar, Y.~Zhu, and L.~Fan, ``Vima: General robot manipulation with multimodal prompts,'' \emph{arXiv preprint arXiv:2210.03094}, vol.~2, no.~3, p.~6, 2022.

\bibitem{yokoyama2024vlfm}
N.~Yokoyama, S.~Ha, D.~Batra, J.~Wang, and B.~Bucher, ``Vlfm: Vision-language frontier maps for zero-shot semantic navigation,'' in \emph{2024 IEEE International Conference on Robotics and Automation (ICRA)}.\hskip 1em plus 0.5em minus 0.4em\relax IEEE, 2024, pp. 42--48.

\bibitem{nasiriany2019planning}
S.~Nasiriany, V.~Pong, S.~Lin, and S.~Levine, ``Planning with goal-conditioned policies,'' \emph{Advances in neural information processing systems}, vol.~32, 2019.

\bibitem{dezfouli2013actions}
A.~Dezfouli and B.~W. Balleine, ``Actions, action sequences and habits: evidence that goal-directed and habitual action control are hierarchically organized,'' \emph{PLoS computational biology}, vol.~9, no.~12, p. e1003364, 2013.

\bibitem{brooks2023instructpix2pix}
T.~Brooks, A.~Holynski, and A.~A. Efros, ``Instructpix2pix: Learning to follow image editing instructions,'' in \emph{Proceedings of the IEEE/CVF Conference on Computer Vision and Pattern Recognition}, 2023, pp. 18\,392--18\,402.

\bibitem{todorov2012mujoco}
E.~Todorov, T.~Erez, and Y.~Tassa, ``Mujoco: A physics engine for model-based control,'' in \emph{2012 IEEE/RSJ International Conference on Intelligent Robots and Systems}.\hskip 1em plus 0.5em minus 0.4em\relax IEEE, 2012, pp. 5026--5033.

\bibitem{lum2024dextrah}
T.~G.~W. Lum, M.~Matak, V.~Makoviychuk, A.~Handa, A.~Allshire, T.~Hermans, N.~D. Ratliff, and K.~Van~Wyk, ``Dextrah-g: Pixels-to-action dexterous arm-hand grasping with geometric fabrics,'' \emph{arXiv preprint arXiv:2407.02274}, 2024.

\bibitem{zhao2023learning}
T.~Z. Zhao, V.~Kumar, S.~Levine, and C.~Finn, ``Learning fine-grained bimanual manipulation with low-cost hardware,'' \emph{arXiv preprint arXiv:2304.13705}, 2023.

\bibitem{chi2023diffusion}
C.~Chi, S.~Feng, Y.~Du, Z.~Xu, E.~Cousineau, B.~Burchfiel, and S.~Song, ``Diffusion policy: Visuomotor policy learning via action diffusion,'' \emph{arXiv preprint arXiv:2303.04137}, 2023.

\bibitem{buamanee2024bi}
T.~Buamanee, M.~Kobayashi, Y.~Uranishi, and H.~Takemura, ``Bi-act: Bilateral control-based imitation learning via action chunking with transformer,'' \emph{arXiv preprint arXiv:2401.17698}, 2024.

\bibitem{mishra2023generative}
U.~A. Mishra, S.~Xue, Y.~Chen, and D.~Xu, ``Generative skill chaining: Long-horizon skill planning with diffusion models,'' in \emph{Conference on Robot Learning}.\hskip 1em plus 0.5em minus 0.4em\relax PMLR, 2023, pp. 2905--2925.

\bibitem{fu2024mobile}
Z.~Fu, T.~Z. Zhao, and C.~Finn, ``Mobile aloha: Learning bimanual mobile manipulation with low-cost whole-body teleoperation,'' \emph{arXiv preprint arXiv:2401.02117}, 2024.

\bibitem{padalkar2023open}
A.~Padalkar, A.~Pooley, A.~Jain, A.~Bewley, A.~Herzog, A.~Irpan, A.~Khazatsky, A.~Rai, A.~Singh, A.~Brohan \emph{et~al.}, ``Open x-embodiment: Robotic learning datasets and rt-x models,'' \emph{arXiv preprint arXiv:2310.08864}, 2023.

\bibitem{ha2023scaling}
H.~Ha, P.~Florence, and S.~Song, ``Scaling up and distilling down: Language-guided robot skill acquisition,'' in \emph{Conference on Robot Learning}.\hskip 1em plus 0.5em minus 0.4em\relax PMLR, 2023, pp. 3766--3777.

\bibitem{reuss2023goal}
M.~Reuss, M.~Li, X.~Jia, and R.~Lioutikov, ``Goal-conditioned imitation learning using score-based diffusion policies,'' \emph{arXiv preprint arXiv:2304.02532}, 2023.

\bibitem{hausman2017multi}
K.~Hausman, Y.~Chebotar, S.~Schaal, G.~Sukhatme, and J.~J. Lim, ``Multi-modal imitation learning from unstructured demonstrations using generative adversarial nets,'' \emph{Advances in neural information processing systems}, vol.~30, 2017.

\bibitem{shridhar2022cliport}
M.~Shridhar, L.~Manuelli, and D.~Fox, ``Cliport: What and where pathways for robotic manipulation,'' in \emph{Conference on robot learning}.\hskip 1em plus 0.5em minus 0.4em\relax PMLR, 2022, pp. 894--906.

\bibitem{brohan2023can}
A.~Brohan, Y.~Chebotar, C.~Finn, K.~Hausman, A.~Herzog, D.~Ho, J.~Ibarz, A.~Irpan, E.~Jang, R.~Julian \emph{et~al.}, ``Do as i can, not as i say: Grounding language in robotic affordances,'' in \emph{Conference on robot learning}.\hskip 1em plus 0.5em minus 0.4em\relax PMLR, 2023, pp. 287--318.

\bibitem{wang2024rise}
C.~Wang, H.~Fang, H.-S. Fang, and C.~Lu, ``Rise: 3d perception makes real-world robot imitation simple and effective,'' \emph{arXiv preprint arXiv:2404.12281}, 2024.

\bibitem{mandlekar2023mimicgen}
A.~Mandlekar, S.~Nasiriany, B.~Wen, I.~Akinola, Y.~Narang, L.~Fan, Y.~Zhu, and D.~Fox, ``Mimicgen: A data generation system for scalable robot learning using human demonstrations,'' \emph{arXiv preprint arXiv:2310.17596}, 2023.

\bibitem{fang2024egocentric}
Z.~Fang, M.~Yang, W.~Zeng, B.~Li, J.~Yue, Z.~Ding, X.~Li, and Z.~Lu, ``Egocentric vision language planning,'' \emph{arXiv preprint arXiv:2408.05802}, 2024.

\bibitem{black2023zero}
K.~Black, M.~Nakamoto, P.~Atreya, H.~Walke, C.~Finn, A.~Kumar, and S.~Levine, ``Zero-shot robotic manipulation with pretrained image-editing diffusion models,'' \emph{arXiv preprint arXiv:2310.10639}, 2023.

\bibitem{reed2022generalist}
S.~Reed, K.~Zolna, E.~Parisotto, S.~G. Colmenarejo, A.~Novikov, G.~Barth-Maron, M.~Gimenez, Y.~Sulsky, J.~Kay, J.~T. Springenberg \emph{et~al.}, ``A generalist agent,'' \emph{arXiv preprint arXiv:2205.06175}, 2022.

\bibitem{alayrac2022flamingo}
J.-B. Alayrac, J.~Donahue, P.~Luc, A.~Miech, I.~Barr, Y.~Hasson, K.~Lenc, A.~Mensch, K.~Millican, M.~Reynolds \emph{et~al.}, ``Flamingo: a visual language model for few-shot learning,'' \emph{Advances in neural information processing systems}, vol.~35, pp. 23\,716--23\,736, 2022.

\bibitem{huang2022language}
W.~Huang, P.~Abbeel, D.~Pathak, and I.~Mordatch, ``Language models as zero-shot planners: Extracting actionable knowledge for embodied agents,'' in \emph{International conference on machine learning}.\hskip 1em plus 0.5em minus 0.4em\relax PMLR, 2022, pp. 9118--9147.

\bibitem{newcombe2015dynamicfusion}
R.~A. Newcombe, D.~Fox, and S.~M. Seitz, ``Dynamicfusion: Reconstruction and tracking of non-rigid scenes in real-time,'' in \emph{Proceedings of the IEEE conference on computer vision and pattern recognition}, 2015, pp. 343--352.

\bibitem{pang2022grasp}
J.~Pang, M.~A. Lodhi, and D.~Tian, ``Grasp-net: Geometric residual analysis and synthesis for point cloud compression,'' in \emph{Proceedings of the 1st International Workshop on Advances in Point Cloud Compression, Processing and Analysis}, 2022, pp. 11--19.

\bibitem{huang2023voxposer}
W.~Huang, C.~Wang, R.~Zhang, Y.~Li, J.~Wu, and L.~Fei-Fei, ``Voxposer: Composable 3d value maps for robotic manipulation with language models,'' \emph{arXiv preprint arXiv:2307.05973}, 2023.

\bibitem{qi2020learning}
W.~Qi, R.~T. Mullapudi, S.~Gupta, and D.~Ramanan, ``Learning to move with affordance maps,'' \emph{arXiv preprint arXiv:2001.02364}, 2020.

\bibitem{yang2023dawn}
Z.~Yang, L.~Li, K.~Lin, J.~Wang, C.-C. Lin, Z.~Liu, and L.~Wang, ``The dawn of lmms: Preliminary explorations with gpt-4v (ision),'' \emph{arXiv preprint arXiv:2309.17421}, vol.~9, no.~1, p.~1, 2023.

\bibitem{zhi2024closed}
P.~Zhi, Z.~Zhang, M.~Han, Z.~Zhang, Z.~Li, Z.~Jiao, B.~Jia, and S.~Huang, ``Closed-loop open-vocabulary mobile manipulation with gpt-4v,'' \emph{arXiv preprint arXiv:2404.10220}, 2024.

\bibitem{zhang2024magicbrush}
K.~Zhang, L.~Mo, W.~Chen, H.~Sun, and Y.~Su, ``Magicbrush: A manually annotated dataset for instruction-guided image editing,'' \emph{Advances in Neural Information Processing Systems}, vol.~36, 2024.

\bibitem{bharadhwaj2023visual}
H.~Bharadhwaj, A.~Gupta, and S.~Tulsiani, ``Visual affordance prediction for guiding robot exploration,'' in \emph{2023 IEEE International Conference on Robotics and Automation (ICRA)}.\hskip 1em plus 0.5em minus 0.4em\relax IEEE, 2023, pp. 3029--3036.

\bibitem{wang2018high}
T.-C. Wang, M.-Y. Liu, J.-Y. Zhu, A.~Tao, J.~Kautz, and B.~Catanzaro, ``High-resolution image synthesis and semantic manipulation with conditional gans,'' in \emph{Proceedings of the IEEE conference on computer vision and pattern recognition}, 2018, pp. 8798--8807.

\bibitem{sridhar2024nomad}
A.~Sridhar, D.~Shah, C.~Glossop, and S.~Levine, ``Nomad: Goal masked diffusion policies for navigation and exploration,'' in \emph{2024 IEEE International Conference on Robotics and Automation (ICRA)}.\hskip 1em plus 0.5em minus 0.4em\relax IEEE, 2024, pp. 63--70.

\bibitem{vaswani2017attention}
A.~Vaswani, ``Attention is all you need,'' \emph{Advances in Neural Information Processing Systems}, 2017.

\bibitem{kingma2013auto}
D.~P. Kingma, ``Auto-encoding variational bayes,'' \emph{arXiv preprint arXiv:1312.6114}, 2013.

\bibitem{sohn2015learning}
K.~Sohn, H.~Lee, and X.~Yan, ``Learning structured output representation using deep conditional generative models,'' \emph{Advances in neural information processing systems}, vol.~28, 2015.

\bibitem{gadre2023cows}
S.~Y. Gadre, M.~Wortsman, G.~Ilharco, L.~Schmidt, and S.~Song, ``Cows on pasture: Baselines and benchmarks for language-driven zero-shot object navigation,'' in \emph{Proceedings of the IEEE/CVF Conference on Computer Vision and Pattern Recognition}, 2023, pp. 23\,171--23\,181.

\bibitem{shi2024yell}
L.~X. Shi, Z.~Hu, T.~Z. Zhao, A.~Sharma, K.~Pertsch, J.~Luo, S.~Levine, and C.~Finn, ``Yell at your robot: Improving on-the-fly from language corrections,'' \emph{arXiv preprint arXiv:2403.12910}, 2024.

\bibitem{he2016deep}
K.~He, X.~Zhang, S.~Ren, and J.~Sun, ``Deep residual learning for image recognition,'' in \emph{Proceedings of the IEEE conference on computer vision and pattern recognition}, 2016, pp. 770--778.

\bibitem{perez2018film}
E.~Perez, F.~Strub, H.~De~Vries, V.~Dumoulin, and A.~Courville, ``Film: Visual reasoning with a general conditioning layer,'' in \emph{Proceedings of the AAAI conference on artificial intelligence}, vol.~32, no.~1, 2018.

\bibitem{james2020rlbench}
S.~James, Z.~Ma, D.~R. Arrojo, and A.~J. Davison, ``Rlbench: The robot learning benchmark \& learning environment,'' \emph{IEEE Robotics and Automation Letters}, vol.~5, no.~2, pp. 3019--3026, 2020.

\bibitem{liu2023libero}
B.~Liu, Y.~Zhu, C.~Gao, Y.~Feng, Q.~Liu, Y.~Zhu, and P.~Stone, ``Libero: Benchmarking knowledge transfer for lifelong robot learning,'' \emph{Advances in Neural Information Processing Systems}, vol.~36, pp. 44\,776--44\,791, 2023.

\bibitem{wu2024gellogenerallowcostintuitive}
\BIBentryALTinterwordspacing
P.~Wu, Y.~Shentu, Z.~Yi, X.~Lin, and P.~Abbeel, ``Gello: A general, low-cost, and intuitive teleoperation framework for robot manipulators,'' 2024. [Online]. Available: \url{https://arxiv.org/abs/2309.13037}
\BIBentrySTDinterwordspacing

\bibitem{dasari2023pgdm}
S.~Dasari, A.~Gupta, and V.~Kumar, ``Learning dexterous manipulation from exemplar object trajectories and pre-grasps,'' in \emph{IEEE International Conference on Robotics and Automation 2023}, 2023.

\bibitem{Xiang_2020_SAPIEN}
F.~Xiang, Y.~Qin, K.~Mo, Y.~Xia, H.~Zhu, F.~Liu, M.~Liu, H.~Jiang, Y.~Yuan, H.~Wang, L.~Yi, A.~X. Chang, L.~J. Guibas, and H.~Su, ``{SAPIEN}: A simulated part-based interactive environment,'' in \emph{The IEEE Conference on Computer Vision and Pattern Recognition (CVPR)}, June 2020.

\bibitem{Mo_2019_CVPR}
K.~Mo, S.~Zhu, A.~X. Chang, L.~Yi, S.~Tripathi, L.~J. Guibas, and H.~Su, ``{PartNet}: A large-scale benchmark for fine-grained and hierarchical part-level {3D} object understanding,'' in \emph{The IEEE Conference on Computer Vision and Pattern Recognition (CVPR)}, June 2019.

\bibitem{chang2015shapenet}
A.~X. Chang, T.~Funkhouser, L.~Guibas, P.~Hanrahan, Q.~Huang, Z.~Li, S.~Savarese, M.~Savva, S.~Song, H.~Su \emph{et~al.}, ``Shapenet: An information-rich 3d model repository,'' \emph{arXiv preprint arXiv:1512.03012}, 2015.

\end{thebibliography}

\end{document}